\documentclass[sigconf]{acmart}
\usepackage{multirow}
\usepackage{graphicx}
\usepackage{subfig}
\newcommand{\eg}{\emph{e.g.,}~}
\newcommand{\etal}{\emph{et al.}~}
\newcommand{\ie}{\emph{i.e.,}~}

\usepackage{enumitem}
\setlist[itemize]{leftmargin=*}

\AtBeginDocument{%
  \providecommand\BibTeX{{%
    \normalfont B\kern-0.5em{\scshape i\kern-0.25em b}\kern-0.8em\TeX}}}

\copyrightyear{2020} 
\acmYear{2020} 
\setcopyright{acmcopyright}\acmConference[SIGIR '20]{Proceedings of the 43rd International ACM SIGIR Conference on Research and Development in Information Retrieval}{July 25--30, 2020}{Virtual Event, China}
\acmBooktitle{Proceedings of the 43rd International ACM SIGIR Conference on Research and Development in Information Retrieval (SIGIR '20), July 25--30, 2020, Virtual Event, China}
\acmPrice{15.00}
\acmDOI{10.1145/3397271.3401151}
\acmISBN{978-1-4503-8016-4/20/07}

\begin{document}
	\fancyhead{}

\title{Tree-Augmented Cross-Modal Encoding for Complex-Query Video Retrieval}

\author{Xun~Yang}
\email{xunyang@nus.edu.sg}
\affiliation{%
  \institution{National University of Singapore}
}
\author{Jianfeng~Dong}
\authornote{Corresponding Author.}
\email{dongjf24@gmail.com}
\affiliation{%
 Zhejiang Gongshang University
}
\author{Yixin~Cao}
\email{caoyixin2011@gmail.com}
\affiliation{%
  \institution{National University of Singapore}
}
\author{Xun~Wang}
\email{wx@zjgsu.edu.cn}
\affiliation{%
 Zhejiang Gongshang University
}
\author{Meng~Wang}
\email{wangmeng@hfut.edu.cn}
\affiliation{%
  \institution{Hefei University of Technology}
}
\author{Tat-Seng~Chua}
\email{chuats@comp.nus.edu.sg}
\affiliation{%
  \institution{National University of Singapore}
}

\begin{abstract}
The rapid growth of user-generated videos on the Internet has intensified the need for text-based video retrieval systems. Traditional methods mainly favor the concept-based paradigm on retrieval with simple queries, which are usually ineffective for complex queries that carry far more complex semantics. Recently, embedding-based paradigm has emerged as a popular approach. It aims to map the queries and videos into a shared embedding space where semantically-similar texts and videos are much closer to each other. Despite its simplicity, it forgoes the exploitation of the syntactic structure of text queries, making it suboptimal to model the complex queries.

To facilitate video retrieval with complex queries, we propose a Tree-augmented Cross-modal Encoding method by jointly learning the linguistic structure of queries and the temporal representation of videos. Specifically, given a complex user query, we first recursively compose a latent semantic tree to structurally describe the text query. We then design a tree-augmented query encoder to derive structure-aware query representation and a temporal attentive video encoder to model the temporal characteristics of videos. Finally, both the query and videos are mapped into a joint embedding space for matching and ranking. In this approach, we have a better understanding and modeling of the complex queries, thereby achieving a better video retrieval performance. Extensive experiments on large scale video retrieval benchmark datasets demonstrate the effectiveness of our approach.
\end{abstract}

\ccsdesc[500]{Information systems~Multimedia and multimodal retrieval}
\ccsdesc[500]{Information systems~Video search}

\keywords{Multimedia retrieval, Video Search, Natural Language Understanding, Latent Tree Structure}

\maketitle

\section{Introduction}
With the exponential growth of user-generated videos on the Internet, searching the videos of interest has been an indispensable activity in people's daily lives. Meanwhile, text-based video retrieval has attracted world-wide research interests and achieved promising progress for retrieval with keyword-based simple queries \cite{snoek2009concept}. However, the expression of text query has been transformed from the keyword-based mechanism to complex queries in recent years. A complex query is usually defined as a natural language query, \eg ``\textit{Two girls are laughing together and then another throws her folded laundry around the room}", which carries far more complex semantics than short queries. How to correctly understand the complex queries has become one of the key challenges in the multimedia information retrieval community. 

\begin{figure}[tb!]
	\centering\includegraphics[width=0.95\columnwidth]{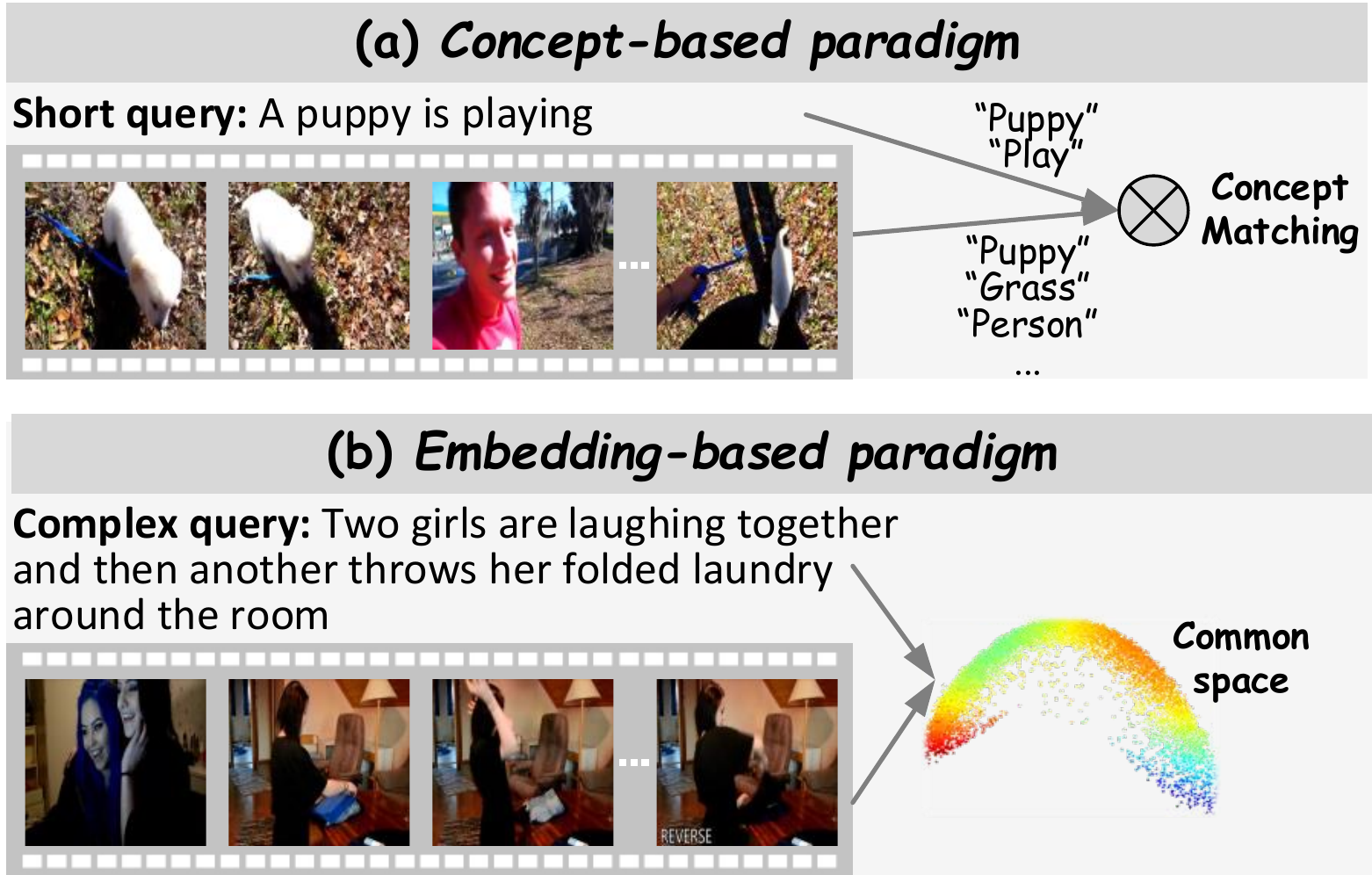}
	\vspace{-0.1in}
	\caption{Concept-based paradigm \textit{vs.} Embedding-based paradigm for text-based video retrieval.}\label{fig1:showcase}
	\vspace{-0.1in}
\end{figure}

Existing efforts on video retrieval with complex queries can be roughly categorized into two groups: 1) \textbf{Concept-based} paradigm \cite{yuan2011learning,yuan2011utilizing,tv16-nii,tv16-certh,icmr2017-certh-avs,tv17-waseda,tv17-vireo}, as shown in Figure \ref{fig1:showcase} (a). It usually uses a large set of visual concepts to describe the video content, then transforms the text query into a set of primitive concepts, and finally performs video retrieval by aggregating the matching results from different concepts \cite{yuan2011utilizing}. Despite its efficiency, it is usually ineffective for complex long queries, since they carry complex linguistic context and cannot be simply treated as an aggregation of extracted concepts. Besides, it is also quite challenging to effectively train concept classifiers and select the relevant concepts. 2) \textbf{Embedding-based} paradigm \cite{dong2019dual,miech2019howto100m,miech2018learning,wray2019fine,yu2018joint,li2019w2vv++,cao2019video}, as shown in Figure \ref{fig1:showcase} (b). Recent efforts proposed to learn a joint text-video embedding space \cite{mithun2018learning,dong2019dual,miech2019howto100m,miech2018learning} to support video retrieval by leveraging the strong representation ability of deep neural networks \cite{nips2012-hinton,sak2014long}. The natural language queries are usually transformed into dense vector representations by Recurrent Neural Networks (RNNs) \cite{sak2014long}  (\eg Long Short-Term Memory (LSTM) and Gated Recurrent Unit (GRU)) that are powerful for modeling sequence data. 
The videos are usually modeled as a temporal aggregation of frame/clip-level features, extracted from pre-trained Convolutional Neural Networks (CNNs) \cite{nips2012-hinton}. Both the queries and videos are mapped into a shared embedding space where semantically-similar videos and text queries are mapped to close points. Although the embedding-based methods have shown much better performance, simply treating queries holistically as one dense vector representations may obfuscate the keywords or phrases that have rich temporal and semantic cues.

Some prior works proposed to transform the complex queries into structured forms, \eg semantic graph \cite{lin2014visual}, to describe the spatial or semantic relations between concepts. However, such solutions usually require the text query to be well annotated with syntactic labels (\eg part of speech (POS) tag) and rely on complex predefined rules to construct the structure of text queries, which make it hard to be applied in a new scenario with different linguistic expression patterns. Although so much efforts have been devoted to complex-query video retrieval, it still remains to be a very challenging task. 

\begin{figure*}[htbp]
	\centering\includegraphics[width=2.05\columnwidth]{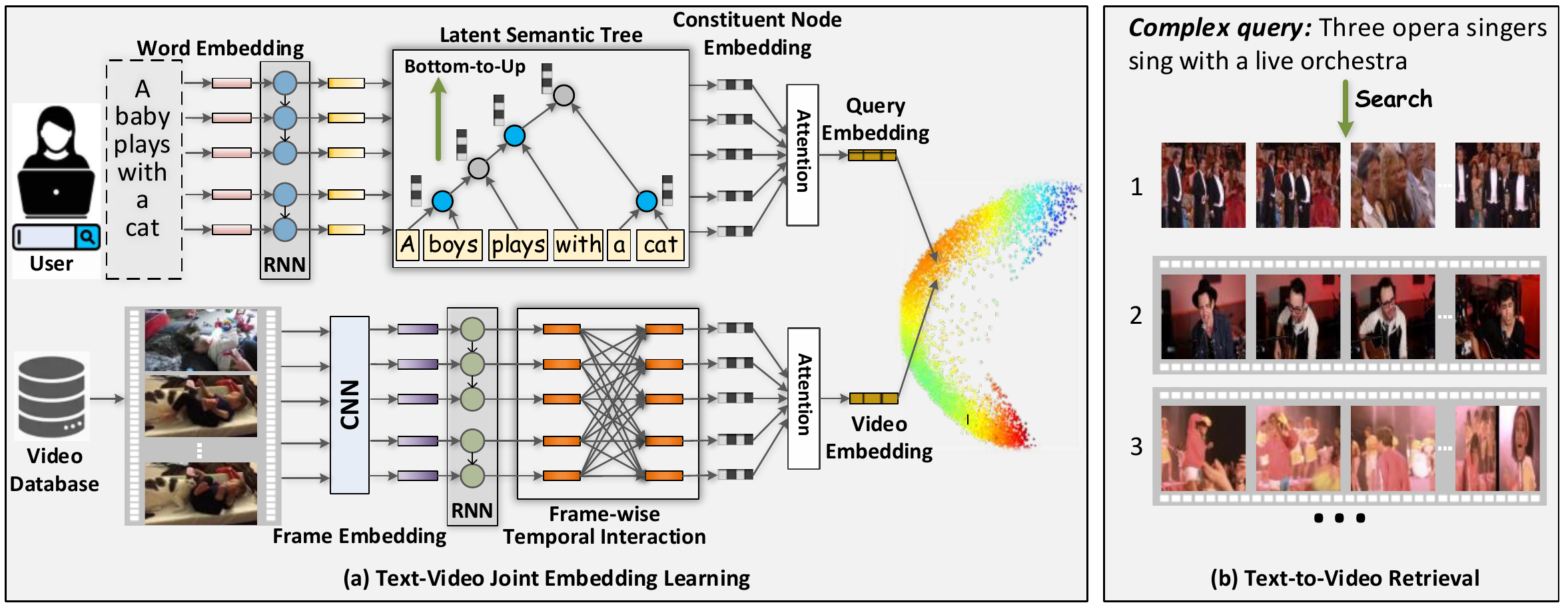}
	\vspace{-0.1in}
	\caption{An illustration of our tree-augmented cross-modal encoding method for complex-query video retrieval.}\label{fig2:framework}
	\vspace{-0.1in}
\end{figure*}

Towards this research goal, this paper aims to model complex queries in a more flexible structure to facilitate the joint learning of the representations of the queries and videos in a unified framework. 
Specifically, we develop a Tree-augmented Cross-modal Encoding (TCE) framework for video retrieval with complex queries. As shown in Figure \ref{fig2:framework}, for the modeling of the complex query, we first recursively compose a Latent Semantic Tree (LST) to describe the query (\eg \textit{A baby plays with a fatty cat}) without any syntactic annotations, where each node (\eg \textit{a baby plays}) denotes a constituent in the complex query. We also propose a memory-augmented node scoring and selection method to inject linguistic context into the construction of LST. We then design a tree-augmented query encoder that identifies the informative constituent nodes in LST and aggregates the constituent embeddings into the structure-aware query representation. 
For the modeling of the videos, we introduce a temporal attentive video encoder that first models the temporal dependence and interaction between frames and then attentively aggregates the frame embeddings into the temporal-attentive video representation. 
Finally, both the user queries and videos are mapped to a text-video joint embedding space where semantically-similar videos and text queries are mapped to close points. All the modules are jointly optimized in an end-to-end fashion using only the paired query-to-video supervision. We evaluate the proposed approach on two large-scale text-to-video retrieval datasets, which clearly demonstrates the effectiveness of each component in our approach. The contributions of this paper are roughly summarized as follows:
\begin{itemize}
    \item We develop a novel complex-query video retrieval framework that can automatically compose a flexible tree structure to model the complex query and derive the query and video representations in a joint text-video embedding space.
    \item We design a memory-augmented node scoring and selection method to explore linguistic context for the tree construction. We also introduce the attention mechanism into the encodings of complex queries and videos, which can identify the informative constituent nodes and frames. 
    \item We conduct extensive experiments on large-scale datasets to demonstrate that our approach can achieve state-of-the-art retrieval performance.
\end{itemize}

\section{Related Work}
In this section, we briefly introduce two representative research directions in text-based video retrieval. One is the concept based methods  and the other one is the embedding based methods. 

Concept based methods \cite{tv16-nii,tv16-certh,icmr2017-certh-avs,tv17-waseda,tv17-vireo} 
mainly rely on establishing cross-modal associations via concepts~\cite{hong2015learning}. Markatopoulou \etal \cite{tv16-certh,icmr2017-certh-avs} first utilized relatively complex linguistic rules to extract relevant concepts from a given query and used pre-trained CNNs to detect the objects and scenes in video frames. Then the similarity between a given query and a specific video is measured by concept matching.
Ueki \etal \cite{tv17-waseda} depended on a much larger concept vocabulary. In addition to pre-trained CNNs, they additionally trained SVM-based classifiers to automatically annotate the videos.
Snoek \etal \cite{tv2017-uvaruc} trained a more elegant model, called VideoStory, from freely available web videos to annotate videos, while they still represented the textual query by selecting concepts based on part-of-speech tagging heuristically. 
Despite the promising performance, the concept based methods still face many challenges, \eg how to specify a set of concepts and how to extract relevant concepts for both textual queries and videos. Moreover, the extraction of concepts from videos and textual queries are usually treated independently, which makes it suboptimal to explore the relations between two modalities.
In contrast, our method is concept free and jointly learns the representation of textual queries and videos.

Deep learning technologies have been popularly explored for video retrieval recently \cite{li2019deep,aaai2015-xu-video,wray2019fine, miech2018learning,miech2019howto100m,li2019w2vv++,dong2019dual,yu2018joint,mithun2018learning}. 
Most works proposed to embed textual queries and videos into a common space, and their similarity is measured in this space by distance metric, \eg cosine distance. 
For textual query embedding,  the word2vec models pre-trained on large-scale text corpora 
are increasingly popular \cite{miech2019howto100m,Dian2018Find,wray2019fine,miech2018learning}. However, they ignored the sequential order in textual queries. To alleviate this, Mithun \etal \cite{mithun2018learning} utilized GRU for modeling the word orders. 
Further, Dong \etal \cite{li2019w2vv++} and Li \etal \cite{li2019w2vv++} jointly employed multiple text embedding strategies including bag-of-words, word2vec, and GRU, to obtain robust query representation. In a follow-up work \cite{dong2019dual}, Dong \etal proposed a multi-level text encoding to capture the global, local, and temporal patterns in the textual queries.
Despite their effectiveness, these methods simply treating queries holistically as one dense vector representations, which may obfuscate the keywords or phrases that have rich semantic cues and are less interpretable than the concept-based paradigm. In this work, we explicitly explore the syntactic structure of natural language query, thus will help to better understand the search intention. Lin \etal \cite{lin2014visual} and Xu \etal \cite{aaai2015-xu-video}  have made attempts in this direction. Lin \etal first obtained the parse tree of the textual query, and modeled the word dependency based on a series of manually derived rules. In \cite{aaai2015-xu-video}, Xu \etal constructed the dependency-tree structure based on subject-verb-object triplets extracted from a sentence and modeled the structure by a recursive neural network. 

For video embedding, a typical approach is to first extract the frame-level features by pre-trained CNNs and subsequently aggregate them into a video representation. To obtain the video-level feature, mean pooling and max pooling are common choices \cite{aaai2015-xu-video,dong2018predicting,mithun2018learning,miech2019howto100m}. Yu \etal \cite{yu2017end} used LSTM to model the temporal information, where frame-level features are sequentially fed into LSTM, and the hidden vector at the last step is used as the video feature. Dong \etal~\cite{dong2019dual} also explicitly exploited the global and local patterns in videos to obtain a multi-level video representation. In this work, we design a temporal attentive video encoder that jointly models the temporal dependence between consecutive frames by RNNs and frame-wise temporal interaction by using the multi-head attention mechanism.

Natural language query-based retrieval techniques have also been successfully applied for domain-specific object retrieval in the filed of video surveillance or E-commerce, such as text-based person search~\cite{li2017person} and its application in person re-identification~\cite{Niu2020TIP,yang2017person,yang2017enhancing} and dialog-based fashion retrieval~\cite{guo2018dialog}. In \cite{li2017person}, Li \etal collected a large-scale person description dataset with detailed natural language annotations and person samples from various sources and proposed a novel recurrent neural network with gated neural attention for person search. Niu \etal ~\cite{Niu2020TIP} designed a multi-granularity image-text alignments model for better modeling the similarity between text description and person images. In~\cite{guo2018dialog}, Guo \etal introduced the reinforcement learning techniques to the task of dialog-based interactive image search that enables users to provide feedback via natural language. 

\section{The proposed Approach}
This paper proposes to tackle the content-based complex-query video retrieval task, in which the query is a natural language sentence that describes a video. 
We basically follow the embedding-based paradigm that embeds the queries and videos into a joint embedding space where texts and videos can be easily matched and ranked. In this section, we first introduce an approach to recursively compose a latent semantic tree to model the complex query in Section 3.1. Then we introduce how to obtain the vector representations of the query and videos in Section 3.2 and 3.3, followed by the joint optimization of the query and video embeddings in the same space in Section 3.4.
\subsection{LST: Latent Semantic Tree}\label{section3.1}
To better understand the complex query, this work proposes to use the Tree-structured LSTM (TreeLSTM) \cite{tai2015improved} to recursively compose a latent semantic tree (LST) in a bottom-to-up fashion to structurally describe each given query. 
Following \cite{choi2018learning}, the LST structure is formulated as a binary recursive tree with two kinds of nodes: \textit{leaf nodes} (\ie words) and \textit{parent nodes} (\ie constituents). A parent node takes in two adjacent child nodes and describes more complex semantics than its child nodes. In this section, we first briefly describe how to apply TreeLSTM to compute the parent node representation from its two child nodes and then describe how to select the parent node at each layer for recursively building the LST. \\
\noindent{\textbf{TreeLSTM}}. Given the representations of two adjacent child nodes ${\left(\mathbf{h}_i,\mathbf{c}_i\right)}$ and ${\left(\mathbf{h}_{i+1},\mathbf{c}_{i+1}\right)}$ as inputs, the parent node representation ${\left(\mathbf{h}_p,\mathbf{c}_p\right)}$ is computed by
    \begin{equation}\label{treelstm1}
\left[\begin{array}{c}
\mathbf{i}\\
\mathbf{f}_l\\
\mathbf{f}_r\\
\mathbf{o}\\
\mathbf{g}
\end{array} 
\right]=  
\left[\begin{array}{c}
\mathbf{\sigma}\\
\mathbf{\sigma}\\
\mathbf{\sigma}\\
\mathbf{\sigma}\\
\mathbf{\mathrm{tanh}}
\end{array} 
\right] \left(\mathbf{W}^p \left[\begin{array}{c}
\mathbf{h}_{i}\\
\mathbf{h}_{i+1}
\end{array} 
\right] + \mathbf{b}^p\right),    
\end{equation}
\begin{equation}\label{treelstm2}
\mathbf{{c}}_p  = \mathbf{f}_l\odot \mathbf{c}_{i} + \mathbf{f}_r\odot \mathbf{c}_{i+1} + \mathbf{i}\odot \mathbf{g},
\end{equation}
\begin{equation}\label{treelstm3}
\mathbf{{h}}_p = \mathbf{o}\odot \mathrm{tanh}(\mathbf{{c}}_p),
\end{equation}    
where $\mathbf{W}^p\in\mathbb{R}^{{5d_t}\times{2d_t}}$ and $\mathbf{b}^p\in\mathbb{R}^{5d_t}$ are trainable parameters, $\sigma(\cdot)$ denotes the activation function \textit{sigmoid}, and $\odot$ denotes the element-wise product. Similar to the standard LSTM, each node is represented by a hidden state  $\mathbf{h}\in\mathbb{R}^{d_t}$ and a cell state  $\mathbf{c}\in\mathbb{R}^{d_t}$.\\
\noindent{\textbf{Layer-wise Node Transformation}}. At the bottom layer, given a query with $N$ words as inputs, we first represent it as a sequence of word embeddings and then transform the word embeddings to the representations of leaf nodes at the corresponding locations. Assume the $t$-th layer of the LST consists of $N_t$ nodes $\{\mathbf{r}_i^t={(\mathbf{h}^t_i,\mathbf{c}^t_i)}\}^{N_t}_{i=1}$. If two adjacent nodes $\mathbf{r}_i^t$ and $\mathbf{r}_{i+1}^t$ are selected to be merged, then we can utilize the above-mentioned TreeLSTM to compute the representation of the parent node  $\mathbf{r}_i^{t+1}=\textrm{TreeLSTM}(\mathbf{r}_i^t, \mathbf{r}_{i+1}^t)$ at the $t+1$ layer. The representations of the unselected nodes are directly copied to the corresponding positions at the $t+1$ layer. 

\noindent{\textbf{Memory-augmented Node Scoring and Selection}}. The key step to the building of LST is how to accurately select the parent node at each layer. Previous work \cite{choi2018learning} proposed to enumerate all adjacent two nodes (\eg $\mathbf{r}^{t}_i$ and $\mathbf{r}^{t}_{i+1}$ ) to compose the parent node candidates $\{\mathbf{r}^{t+1}_{i}\}^{N_{t+1}}_{i=1}$ and compute their representations by feeding the two consecutive child nodes into the TreeLSTM, and then select the best parent node candidate based on a node scoring module. Choi \etal \cite{choi2018learning} implemented the scoring module by first introducing a global \textit{query vector} and then computing the inner-product between the query vector and the hidden states of parent node candidates, followed by a softmax operation. Despite its simplicity, it is still difficult to effectively decide the best candidate, due to the ambiguity of language and the limited capacity of hidden state \cite{collins2016capacity,yogatama2018memory} to remember the input history, especially when the given query is very long. To address this issue,  we design a memory-augmented scoring module $ f_{score}\left(\cdot;\Theta_{score}\right)$ to select the parent node:
 \begin{equation}
  s^{t+1}_i = f_{score}\left(\mathbf{{r}}^{t+1}_i,\mathbf{u}^{t+1}_i;\Theta_{score}\right),
 \end{equation}
 where $s^{t+1}_i$ denotes the probability of the $i$-th parent node candidate being selected and $\mathbf{u}^{t+1}_i$ is a node-specific context vector derived from a \textbf{global memory} $\mathbf{M}$ which stores the semantic context. The global memory is defined as the set of leaf node hidden state representations $\mathbf{M}=[\mathbf{h}^1_1, \mathbf{h}^1_2,\cdots,\mathbf{h}^1_N]\in\mathbb{R}^{N\times {d_t}}$ at the bottom layer which preserves the original semantic context in the given sentence.
 To obtain the context vector $\mathbf{u}^{t+1}_i$, we use the hidden state $\mathbf{h}^{t+1}_i$ of each parent node candidate to query the global memory and then attentively aggregate the global memory $\mathbf{M}$:
 \begin{equation}
a^{t+1}_{ij} = \textrm{Softmax}\left( ( \mathbf{{h}}^{t+1}_i  )^{\textrm{T}}  \sigma( \mathbf{W}_m  \mathbf{h}^1_j + \mathbf{b}_m)/\sqrt{d_t}\right), \mathbf{u}^{t+1}_i = (\mathbf{a}^{t+1}_i)^{\textrm{T}}\mathbf{M},
 \end{equation}
where $\mathbf{a}^{t+1}_{i}=[a^{t+1}_{i1},a^{t+1}_{i2},\cdots,a^{t+1}_{iN}]$ is  the normalized attention vector over the memory and $\mathbf{W}_m\in\mathbb{R}^{d_t \times {d_t}}$ and $\mathbf{b}_m\in\mathbb{R}^{{d_t}}$ are trainable parameters. We then implement our scoring module as: 
  \begin{equation}\label{score}
  s^{t+1}_i=  \textrm{Softmax} \left(\mathbf{w}_s^{\textrm{T}} \sigma\Big(\mathbf{W}_s\left[\begin{array}{c}
  \mathbf{h}_{i}^{t+1}\\
  \mathbf{u}_{i}^{t+1}
  \end{array} 
  \right] + \mathbf{b}_s\Big)/\sqrt{2d_t}
   \right),
 \end{equation}
where $\sigma(\cdot)$ is the nonlinear activation function $ \mathrm{ReLU}$ and $\mathbf{w}_s\in\mathbb{R}^{2d_t}$, $\mathbf{b}_s\in\mathbb{R}^{2d_t}$, and $\mathbf{W}_s\in\mathbb{R}^{2d_t \times {2d_t}}$ are trainable parameters. The main intuition of this node scoring module is to inject the semantic context into each decision for a better parent node selection. In such a recursive process, we select the candidate with the maximum validity score using Eq. (\ref{score}) based on the Straight-Through (ST) Gumberl-Softmax estimator \cite{gu2018neural}. In the forward pass, the ST Gumberl-Softmax estimator discretizes the continuous signal, while in the backward pass, the continuous signals are used for stable training. Note that only the representation of the selected node is updated using the outputs of Eq. (\ref{treelstm1}), (\ref{treelstm2}), and (\ref{treelstm3}). The other nodes that are not selected are not updated. 

 The above procedure is recursively repeated until only a single node is left. By this procedure, we can automatically compose a $N$-layers binary latent semantic tree with semantically-meaningful constituents to better understand the complex query without any syntactic annotations.

\subsection{Tree-augmented Query Encoder}\label{queryEncoder}
 \noindent{\textbf{One-hot Representation}}. Given a query $\mathcal{Q}=\{q_1, q_2, \cdots, q_{N}\}$, we first represent it as a sequence of one-hot vectors $\{\mathbf{q}'_1, \mathbf{q}'_2, \cdots, \mathbf{q}'_{N}\}$, where $\mathbf{q}'_t$ indicates the vector of the t-th word. We further convert the word vectors to word dense representations $\{\mathbf{q}_1, \mathbf{q}_2, \cdots, \mathbf{q}_{N}\}$ based on pretrained word embedding matrix~\cite{word2vec}.
 
  \noindent{\textbf{Leaf Node LSTM}}. We use RNNs as the basic sequence modeling block. For keeping consistency with the TreeLSTM in our LST module, we use LSTM to transform the word embeddings to the leaf node representations at the bottom layer. More formally, the LSTM unit, at the $i$-th time step, takes the features of the current word $\mathbf{q}_i$, previous hidden state $\mathbf{h}^1_{i-1}$, and cell state $\mathbf{c}^1_{i-1}$ as inputs, and yields the current hidden state $\mathbf{h}^1_i$ and cell state $\mathbf{c}^1_i$:
 \begin{equation}\label{leafLSTM}
 	\big(\mathbf{h}^1_i, \mathbf{c}^1_i\big) = \textrm{LSTM}\big(\mathbf{q}_i, \mathbf{h}^1_{i-1}, \mathbf{c}^1_{i-1}\big).
 \end{equation}
Eq. (\ref{leafLSTM}) functions as the leaf node transformation module.\\ 
\noindent{\textbf{Tree Construction}}. The outputs of Eq. (\ref{leafLSTM}) are directly fed into the TreeLSTM module for the transformation of parent node candidates, as detailed in Eq. (\ref{treelstm1}), (\ref{treelstm2}), and (\ref{treelstm3}). After $N$ steps of the transformation, scoring, and selection, as described in the Section 3.1, we recursively compose a $N$-layers latent semantic tree, consisting of $N$-1 constituent nodes (\ie parent nodes), formulated by 
\begin{equation}
\{\mathbf{e}_1,\mathbf{e}_2,\cdots,\mathbf{e}_{N-1}\}=\textrm{LSTree}(\{\mathbf{q}_1, \mathbf{q}_2, \cdots, \mathbf{q}_{N}\}),
\end{equation}  
where $\textrm{LSTree}$ indicates the overall tree construction procedure and $\mathbf{e}_i\in\mathbb{R}^{d_t}$ denotes the representation of the $i$-th constituent node. 
The tree can clearly describe the syntactic structure of complex queries, which is helpful to better understand the user query. A similar procedure of tree construction can be found in \cite{choi2018learning}.

\noindent{\textbf{Structure-aware Query Representation}}. The next step is to derive the query representation based on the recursively extracted constituent nodes in the LST. In previous work \cite{choi2018learning,shi2019visually}, only the last constituent node is used for task-specific inference. However, as mentioned previously, the complex query usually consists of multiple visual concepts and their reference descriptions, in which some concepts or reference descriptions may not have clear visual evidence or just have very short temporal durations in the videos. The last constituent node may not effectively cover the full linguistic context of the complex queries. In this work, we introduce an attention network to explore the importance of each constituent and then derive the structure-aware query representation by attending to the informative constituent nodes:  
\begin{equation}\label{queryatt}
\beta_i = \mathrm{Softmax}\left(\mathbf{u}^{\textrm{T}}_{ta} \sigma\big( \mathbf{W}_{ta} \mathbf{e}_i + \mathbf{b}_{ta}\big)/\sqrt{d_{ta}} \right), \quad \mathbf{\bar{q}} = \sum_{i=1}^{N}\beta_i \mathbf{e}_i,
\end{equation}
where $\sigma(\cdot)$ is the non-linear activation function ReLU and $\mathbf{W}_{ta}\in{\mathbb{R}^{d_{ta} \times {d_t}}}$, $\mathbf{b}_{ta} \in{\mathbb{R}^{d_{ta}}}$, and $\mathbf{u}_{ta} \in{\mathbb{R}^{d_{ta}}}$ are trainable parameters, $\beta_i $ denotes the normalized importance score of the node $\mathbf{e}_i$, and $\mathbf{\bar{q}}\in{\mathbb{R}^{d_t}}$ denotes the query representation that aggregates the representations of all constituent nodes.

\subsection{Temporal-Attentive Video Encoder}\label{vidEncoder}
Given a video clip $\mathcal{V}$, we first sample uniformly a sequence of video frames $\{v_1, v_2, \cdots, v_M\}$ from $\mathcal{V}$ with a pre-specified interval. We extract the frame features using pre-trained CNNs and represent the video clip as $\mathbf{V}=\{\mathbf{v}_t\}_{t=1}^M$ where $\mathbf{v}_t\in\mathbb{R}^{d^*_v}$ denotes the frame vector of the $t$-th frame. In this paper, we deal with two types of video characteristics: 1) temporal dependence between consecutive frames along the sequence, and 2) frame-wise temporal interaction over the whole video space.

\noindent{\textbf{Temporal Dependence Modeling}}. We leverage the GRU to model the temporal dependence between consecutive frames. 
At each time step, GRU takes the feature vector of the current frame and the hidden state of the previous frame as inputs and yields the hidden state of the current frame:
\begin{equation}\label{vidGRU}
\mathbf{h}'_t = \textrm{GRU}\left(\mathbf{v}_t, \mathbf{h}'_{t-1}\right),
\end{equation} 
where $\mathbf{h}'_t \in\mathbb{R}^{d_v}$ denotes the hidden state of the $t$-th frame. By the operation in Eq. (\ref{vidGRU}), we can effectively capture the dependence between adjacent frames. For representing a video (clip), previous works either leverage the last hidden state or aggregate all the hidden states of frames using average-pooling, forgoing modeling the frame interaction over the whole video space.

\noindent{\textbf{Frame-wise Temporal Interaction Modeling}}. To further enhance of the video sequence representation, we propose to leverage the frame-wise correlation based on the multi-head self-attention mechanism \cite{vaswani2017attention}. Given a video sequence $\mathbf{V}=\{\mathbf{h}'_t\}_{t=1}^M$ produced by the GRU operation in Eq. (\ref{vidGRU}), the basic idea is to first project the video sequence representation $\mathbf{V}$ into multiple embedding spaces and perform scaled dot-product attention between \textit{query} frame and \textit{key} frame, followed by a softmax operation to obtain the normalized weights on the \textit{value} frames. We finally concatenate the outputs from multiple attention spaces as the final \textit{value}:
\begin{equation}
\mathbf{\hat{V}}^i = \mathrm{Softmax}\bigg(\frac{1}{\sqrt{d_i}}{\left(\mathbf{W}^i_Q \mathbf{V}\right)^{\textrm{T}} \mathbf{W}^i_K \mathbf{V}} \bigg) \mathbf{W}^i_V \mathbf{V},
\end{equation}
\begin{equation}
\mathbf{\hat{V}} =   {Norm}\left(\mathbf{V}+\mathbf{W}^p\left({Concat}\Big(\mathbf{\hat{V}}^1,\mathbf{\hat{V}}^2,\cdots,\mathbf{\hat{V}}^Z\Big)\right)\right),
\end{equation}
where $\mathbf{W}^i_Q\in\mathbb{R}^{d_i\times {d_v}}$, $\mathbf{W}^i_K\in\mathbb{R}^{d_i\times {d_v}}$, and $\mathbf{W}^i_V\in\mathbb{R}^{d_i\times {d_v}}$ are three trainable parameters that transform the original input $\mathbf{V}$ to the \textit{query}, \textit{key}, and \textit{value} matrices in the $i$-th attention space with dimension $d_i$. $\mathbf{\hat{V}}^i \in\mathbb{R}^{d_i\times M}$ denotes the attended \textit{value} in the $i$-th attention space. ${Concat(\cdot)}$ denotes the concatenation operation.   $\mathbf{W}^p\in\mathbb{R}^{d_v\times {d_v}}$ is a trainable parameter that projects the concatenated features into original space. ${Norm(\cdot)}$ denotes the LayerNorm operation. $\mathbf{\hat{V}} =\{\mathbf{\hat{v}}_t  \in\mathbb{R}^{d_v}  \}_{t=1}^M $  is the final video sequence representation. The above multi-head attention mechanism allows the model to jointly attend to information from different representation spaces at different positions, which effectively captures the feature interaction among frames. 

\noindent{\textbf{Temporal-attentive Video Representation}}. To make informative frames (\eg foreground frames) contribute more to the final video representation, we design a temporal attention neural network with three trainable parameters $\mathbf{u}_{va}\in\mathbb{R}^{d_{va}}$, $\mathbf{b}_{va}\in\mathbb{R}^{d_{va}}$, and $\mathbf{W}_{va}\in\mathbb{R}^{d_{va} \times {d_v}}$:
\begin{equation}\label{vidatt}
\eta_t = \mathrm{Softmax}\left(\mathbf{u}^{\textrm{T}}_{va} \sigma\big( \mathbf{W}_{va} \mathbf{\hat{v}}_t + \mathbf{b}_{va}\big)/\sqrt{d_{va}} \right), \  \mathbf{\bar{v}} = \sum_{t=1}^{M}\eta_t \mathbf{\hat{v}}_t,
\end{equation}
where $\eta_t$ denotes the normalized importance score of the $t$-th frame, and $\mathbf{\bar{v}}\in\mathbb{R}^{d_v}$ denotes the final video representation.
Eq. (\ref{vidatt}) has a similar formulation as Eq. (\ref{queryatt}), both of which are easy to implement and effective to exploit the informative frame/word features for representation. 

\subsection{Text-Video Joint Embedding}\label{jointembedding}
Formally, given a natural language query $\mathcal{Q}=\{q_1, q_2, \cdots, q_{N}\}$ and a video sequence $\mathcal{V}= \{v_1, v_2, \cdots, v_M\}$,
we transform $\mathcal{Q}$ and $\mathcal{V}$ to low-dimensional vector representations $\mathbf{\bar{q}}\in\mathbb{R}^{d_t}$ and $\mathbf{\bar{v}}\in\mathbb{R}^{d_v}$ using the query encoder described in Section \ref{queryEncoder} and  the video encoder in Section \ref{vidEncoder}, respectively. 
Then we map the text query and video into a joint embedding space by two linear projection matrices: $f^t: \mathbb{R}^{d_t} \rightarrow \mathbb{R}^{d^*}$ and $f^v: \mathbb{R}^{d_v} \rightarrow \mathbb{R}^{d^*}$, where we define the cross modal matching score as the cosine similarity:
\begin{equation}\label{cosinesimilarity}
s\left(\mathcal{Q}, \mathcal{V}\right)=\frac{ {f^t(\mathbf{\bar{q}})}^{\textrm{T}} f^v(\mathbf{\bar{v}}) }{\lVert f^t(\mathbf{\bar{q}})  \rVert_2   \lVert f^v(\mathbf{\bar{v}})  \rVert_2  },
\end{equation}
where ${f^t(\mathbf{\bar{q}})}$ and $f^v(\mathbf{\bar{v}})$ are implemented by
\begin{equation}\label{twoMapping}
f^t(\mathbf{\bar{q}}) = \mathbf{W}^*_t \mathbf{\bar{q}} + \mathbf{b}^*_t, \quad f^v(\mathbf{\bar{v}}) = \mathbf{W}^*_v \mathbf{\bar{v}} + \mathbf{b}^*_v,
\end{equation}
where $\mathbf{W}^*_t\in\mathbb{R}^{d^*\times d^{t}}$, $\mathbf{b}^*_t \in \mathbb{R}^{d^*}$, $\mathbf{W}^*_v\in\mathbb{R}^{d^*\times d^{v}}$, and $\mathbf{b}^*_v \in \mathbb{R}^{d^*}$ are trainable parameters.
We expect Eq. (\ref{cosinesimilarity}) to yield a higher score when the video $\mathcal{V}$ is matched with the complex query $\mathcal{Q}$ or a lower score if not match. We also apply a batch normalization \cite{ioffe2015batch} followed by a non-linear activation $\mathrm{Tanh}(\cdot)$ on ${f^t(\mathbf{\bar{q}})}$ and $f^v(\mathbf{\bar{v}})$, respectively, for stable training. Note that both $f^t(\cdot)$ and $f^v(\cdot)$ are not indispensable if we enforce the output of the query encoder to have the same dimension as the output of video encoder, i.e., $d_t=d_v$. We introduce $f^t(\cdot)$ and $f^v(\cdot)$ in Eq. (\ref{cosinesimilarity}) just for more formal expression and also make the section \ref{jointembedding} self-contained. Besides, with $f^t(\cdot)$ and $f^v(\cdot)$,  we can derive much lower dimensional embeddings for fast retrieval without modifying the parameters in the two encoders.\\
\noindent{\textbf{Loss Function}}: To train the model, we use the margin ranking loss to optimize the network with a batch-hard negative sampling strategy. More formally, during training, we sample a batch of query-video pair $\mathcal{X}=\{(\mathcal{Q}_i, \mathcal{V}_i)\}_{i=1}^B$. We wish to enforce that, for any given $(\mathcal{Q}_i, \mathcal{V}_i)$, the similarity score $s\left(\mathcal{Q}_i, \mathcal{V}_i\right)$ between a query $\mathcal{Q}_i$ and its ground truth video $\mathcal{V}_i$ is larger than the score of any negative pairs
$s\left(\mathcal{Q}_i, \mathcal{V}_j\right)$ by a large margin, when video $\mathcal{V}_j$ does not match with  query $\mathcal{Q}_i$. The loss on the batch is defined as
\begin{equation}\label{loss}
L\left(\mathcal{X}\right) = \frac{1}{|\mathcal{N}^h|}\sum_{i=1}^B\sum_{j\in\mathcal{N}^h} \!  \mathrm{max}\left( 0, \delta + s\left(\mathcal{Q}_i, \mathcal{V}_j\right) - s\left(\mathcal{Q}_i, \mathcal{V}_i\right)  \right),
\end{equation}
where $\delta$ is the margin ($\delta\in(0,1)$ ). $|\mathcal{N}^h|$ denotes the number of hard negative videos in the set $\mathcal{N}^h$. We found that the hardest negative sample may result in unstable training especially in a large batch, but averaging the costs on all negative samples in a batch will result in slow training. Therefore, in this work, we use a trade-off strategy: we just take into consideration the top $|\mathcal{N}^h|$ negative samples (\eg 5) and average the costs for stable and efficient training. 

\section{Experiment}
A key contribution of this work is to develop a new complex-query video retrieval approach with a tree-augmented cross-modal encoding method. We aim to answer the following research questions via extensive experiments:
(1) \textbf{R1:} How does the proposed method perform compared with state-of-the-art methods?
(2) \textbf{R2:} What are the impacts of different components on the overall performance of our approach?
(3) \textbf{R3:} How does the proposed method perform on different types of complex queries (\eg different lengths and different categories)?  Can the latent semantic tree help to better understand the complex query and drive stronger query representation? 

\subsection{Experimental Settings}
\subsubsection{Datasets} We use two public datasets: MSR-VTT video caption dataset \cite{xu2016msr} and LSMDC movie description dataset \cite{rohrbach2015dataset}.\\ 
\noindent{\textbf{MSR-VTT \cite{xu2016msr}: }} It is an increasingly popular dataset for text-to-video retrieval, consisting of 10K YouTube video clips. Each of them is annotated with 20 crowd-sourced English sentences, which results in a total of 200K unique video-caption pairs. We notice that there are three different dataset partitions for this dataset. The first one is the official partition from \cite{xu2016msr} with 6,513 clips for training, 497 clips for validation, and the remaining 2,990 clips for testing. The second one is from \cite{miech2018learning} with 6,656 clips for training and 1000 test clips for testing. The last one is from \cite{yu2018joint}, 7,010 and 1K video clips are used for training and testing respectively. Note for the last two data partitions, only one sentence associated with each video clip is used as the testing query. For a comprehensive evaluation, we evaluate our proposed model on all data partitions.

\noindent{\textbf{LSMDC \cite{rohrbach2015dataset}: }}It is another popular dataset that contains 118,081 short video clips extracted from 202 movies. Each video clip has only one caption, either extracted from the movie script or from the transcribed audio description. It is originally used for evaluation in the Large Scale Movie Description Challenge (LSMDC). In this work, we only consider the text-to-video task in LSMDC: given a natural language query, the system retrieves the video of interest from the 1,000 test video set.

\begin{table} [tb!]
	\renewcommand{\arraystretch}{1.0}
	\caption{State-of-the-art performance comparison (\%) on MSR-VTT with different dataset splits. 
	Note that TCE uses bidirectional GRU and LSTM for better performance in this experiment based on 1024-D query and video embeddings.}
	\label{tab:sota-msrvtt}
		\vspace{-0.1in}
	\centering 
	\scalebox{0.98}{
		\begin{tabular}{|l|c|c|c|c|}
			\hline
			{\textbf{Method}}  &  \textbf{R@1} & \textbf{R@5} &\textbf{ R@10} & \textbf{MedR}   \\
			\hline\hline
			\textit{\textbf{Data split from}} \cite{xu2016msr}  &  & &  &  \\
			Dong \etal \cite{dong2018predicting}            & 1.8 & 7.0 & 10.9 & 193   \\
			Mithun \etal \cite{mithun2018learning}    & 5.8 & 17.6 & 25.2 & 61  \\
			DualEncoding \cite{dong2019dual}         & 7.7 & 22.0 & 31.8 & 32 \\
			\hline
			TCE                             & \textbf{7.7} & \textbf{22.5} & \textbf{32.1} & \textbf{30} \\
			\hline\hline
			\textit{\textbf{Data split from}} \cite{miech2018learning}  &  & &  &   \\
			Random         & 0.3 & 0.7 & 1.1 & 502  \\
			CCA \cite{wray2019fine}        & 7.0 & 14.4 & 18.7 & 100  \\
			MEE \cite{miech2018learning}      & 12.9 & 36.4 & 51.8 & 10.0  \\
			MMEN (Caption)  \cite{wray2019fine}           & 13.8 & 36.7 & 50.7 & 10.3 \\
			JPoSE \cite{wray2019fine}           & 14.3 & 38.1 & 53.0 &\textbf{9} \\
			\hline
			TCE                             & \textbf{17.1} & \textbf{39.9} & \textbf{53.7} &\textbf{9}\\
			\hline\hline
			\textit{\textbf{Data split from}}  \cite{yu2018joint}   &  & &  &  \\
			Random         & 0.1 & 0.5 & 1.0 & 500  \\
			C+LSTM+SA+FC7 \cite{torabi2016learning}  & 4.2 & 12.9 & 19.9 & 55  \\
			VSE-LSTM \cite{kiros2014unifying}                 & 3.8 & 12.7 & 17.1 & 66 \\
			SNUVL \cite{Yu2016VideoCA}       & 3.5 & 15.9 & 23.8 & 44 \\
			Kaufman \etal  \cite{kaufman2017temporal}  & 4.7 & 16.6 & 24.1 & 41  \\
			CT-SAN   \cite{yu2017end}               & 4.4 & 16.6 & 22.3 & 35  \\
			JSFusion \cite{yu2018joint}               & 10.2 & 31.2 & 43.2 & 13  \\
			Miech \etal  \cite{miech2019howto100m}    & 12.1 & 35.0 & 48.0 & 12 \\
			\hline
			TCE                         & \textbf{16.1} & \textbf{38.0 }&\textbf{ 51.5 }& \textbf{10} \\
			\hline
		\end{tabular}
	}	\vspace{-0.15in}
\end{table}

\subsubsection{Implementation Details}
On MSR-VTT, for the word features, we initialize the word embedding matrix using a 500-D word2vec model provided by \cite{dong2018predicting} which optimized word2vec on English tags of 30 million Flickr images. The textual sequence is fed into a unidirectional LSTM with the hidden size of $d_t$=512 for leaf node transformation. The hidden sizes of the TreeLSTM and query attention modules are set to $d_t$=512 and $d_{ta}$=256, respectively. The final query representation has the dimension of $d_t=$512.
For the video features, we use the frame-level visual features provided by \cite{dong2019dual}, where the 2048-D features are extracted with ResNet-152 \cite{cvpr2016-resnet} pre-trained on ImageNet. The video frame sequences are fed in a unidirectional GRU with the hidden size of $d_v$=512. The output of GRU is further fed into an 8-head attention module. The dimension of each head subspace is 64. The temporal attention module with the hidden size of $d_{va}=$256 aggregates the outputs of multi-head attention module and produces a video representation with the dimension of $d_v$=512. As mentioned previously, since video encoder and query encoder have the same dimension, we omit the two projection matrices in Eq. (\ref{twoMapping}) for compressing the size of parameters. The number of hard negative samples used in Eq. (\ref{loss}) is 5. 

On LSMDC, following \cite{miech2018learning}, we use 300-D \textit{GoogleNews} pre-trained word2vec word embeddings as the input of a unidirectional LSTM with the hidden size of 512, followed by our tree construction and the query attention network using the same setting with MSR-VTT. Since \cite{miech2018learning} did not release the frame-level features, we directly use the provided multi-modal video-level features (appearance, motion,  audios and face) to evaluate the effectiveness of our complex-query modeling module. Note that we do not use the gated embedding module and weighted-fusion of similarity scores in \cite{miech2018learning}. We first transform the multiple-modal features to the default embedding spaces in \cite{miech2018learning} with multiple projection matrices and concatenate the multiple features into a long vector, followed by a feature transformation into the 512-D joint embedding space.

\subsubsection{Evaluation Metrics} Following the setting of \cite{dong2019dual,miech2018learning,yu2018joint}, we report the rank-based performance metrics, namely $\mathrm{R@K}$ ($K = 1, 5, 10$) and Median rank (MedR). $\mathrm{R@K}$ is the percentage of test queries for which at least one relevant item is found among the top-$K$ retrieved results. MedR is the median rank of the first relevant item in the search results. Higher $\mathrm{R@K}$ and lower MedR indicate better performance. 

\subsubsection{Training Details} Our work is implemented using the PyTorch framework. We train our model using the ADAM optimizer and use an initialized learning rate of 0.0005 with a batch size of 128. Each epoch training is just performed using a single GPU and takes no more than 10 minutes.

\subsection{Experimental Results and Analysis}
\begin{table} [tb!]
	\renewcommand{\arraystretch}{1.0}
	\caption{State-of-the-art performance comparison (\%) on LSMDC \cite{rohrbach2015dataset}.  Our TCE performs the best with a much lower-dimensional embedding (512-D). The \textit{Mot.} and \textit{Aud.} refer to the motion feature and audio feature, respectively.}
	\label{tab:sota-MPII}
	\centering 
	\scalebox{0.97}{
		\begin{tabular}{|l|c|c|c|c|}
			\hline
			\textbf{Method}   & \textbf{R@1} & \textbf{R@5} &\textbf{ R@10} & \textbf{MedR}   \\
			\hline\hline
			C+LSTM+SA+FC7 \cite{torabi2016learning}  & 4.3 & 12.6 & 18.9 & 98  \\
			VSE-LSTM \cite{kiros2014unifying}                 & 3.1 & 10.4 & 16.5 & 79  \\
			SNUVL \cite{Yu2016VideoCA}       & 3.6 & 14.7 & 23.9 & 50 \\
			Kaufman \etal \cite{kaufman2017temporal} &4.7 &15.9&23.4&64\\
			CT-SAN \cite{yu2017end}       & 5.1 & 16.3 & 25.2 & 46  \\
			Miech \etal \cite{miech2017learning}  & 7.3 & 19.2 & 27.1 & 52\\
			CCA (FV HGLMM) \cite{klein2015associating}   & 7.5 & 21.7 & 31.0 & 33\\
			JSFusion \cite{yu2018joint}   & 9.1 & 21.2 & 34.1 & 36\\
			Miech \etal. \cite{miech2019howto100m} &7.2 &18.3 &25.0 &44\\
			MEE \cite{miech2018learning}   & 10.2 & 25.0 & 33.1 & \textbf{29}\\			
			\hline\hline
			{TCE (Visual)}   &7.9 & 20.8 & 27.8 & 46  \\
			{TCE (Visual+Mot.)}   &9.7& 23.3 & 34.8 & 32  \\
			{TCE (Visual+Mot.+Aud.)}   &\textbf{10.6 }& \textbf{25.8} & \textbf{35.1} & \textbf{29 } \\
			\hline
	\end{tabular}}
	\vspace{-0.15in}
\end{table}

\subsubsection{Comparison with State-of-the-Arts}
To answer the research question \textbf{R1}, we compare our proposed Tree-augmented Cross-modal Encoding (TCE) with recently proposed state-of-the-art methods:
(1) RNN-based methods: DualEncoding \cite{dong2019dual}, Kaufman \etal  \cite{kaufman2017temporal}, CT-SAN   \cite{yu2017end},  SNUVL \cite{Yu2016VideoCA}, C+LSTM+SA+FC7 \cite{torabi2016learning}, and VSE-LSTM \cite{kiros2014unifying},
(2) Multimodal Fusion methods: Mithun \etal \cite{mithun2018learning} , MEE \cite{miech2018learning}, MMEN \cite{wray2019fine}, and JPoSE \cite{wray2019fine},
 and (3) other methods: JSFusion \cite{yu2018joint},  CCA (FV HGLMM) \cite{klein2015associating}, and Miech \etal \cite{miech2017learning}.
The experimental results on MSR-VTT and LSMDC are summarized, respectively, in Table \ref{tab:sota-msrvtt} and Table \ref{tab:sota-MPII}. Note that there are different dataset splitting strategies of the MSR-VTT dataset. To fairly compare with the reported results of state-of-the-art methods, we  first report our results based on the standard split from the official paper \cite{xu2016msr} and then evaluate our method on the other two splits from \cite{miech2018learning} and 
\cite{yu2018joint}, respectively. 
Unless otherwise stated, we use unidirectional RNNs (512-D) in our experiments by default.

\noindent{\textbf{MSR-VTT: }}Table \ref{tab:sota-msrvtt} clearly shows that our proposed TCE outperforms all other available methods in all three dataset splits. Specifically, on the first split \cite{xu2016msr}, we surpass the results of DualEncoding \cite{dong2019dual} w.r.t. R@5, R@10, and MedR. DualEncoding is the best reported state-of-the-art method on the first split that fuses multi-levels textual and video features for joint embedding learning with the embedding size of 2048. While, our TCE just uses the temporal/sequential features (i.e., the 2nd level features in DualEncoding) with the final embedding size of 1024 for retrieval. Hence, TCE is able to report higher retrieval accuracy while using a much smaller embedding size. TCE also outperforms the multimodal fusion (object, activity, and audio) method in Mithun \etal \cite{mithun2018learning} by a large margin, which indicates the effectiveness of our proposed tree-augmented query modeling and temporal-attentive video sequence modeling methods. Note that our proposed TCE can achieve consistent performance improvement if we integrate some other modalities, like motion features or audio features into the video embedding. We evaluate our method in the multi-modality setting on LSMDC (See Table \ref{tab:sota-MPII}). For the second split \cite{miech2018learning}, we observe a large improvement over the state-of-the-art JPoSE which disentangles the text query into multiple semantic spaces (Verb, Noun) for score-level fusion. Compared with JPoSE, TCE directly composes a latent semantic tree to describe the user complex query in an end-to-end manner and also includes an attention mechanism to capture the most informative constituent nodes in the tree. A similar improvement can also be observed in the third split \cite{yu2018joint}, which further validates the effectiveness of TCE.

\noindent{\textbf{LSMDC: }}Table \ref{tab:sota-MPII} compares the performance of TCE with nearly all reported results on the LSMDC video clip retrieval task. The results again show that our proposed TCE performs the best on this challenging benchmark dataset. Specifically, we outperform the MEE method by a relative improvement of 6\% w.r.t. R@10. We use the same multi-modal video features with MEE, but with a simpler feature fusion strategy, \ie  concatenation. We can observe a more significant improvement over the JSFusion method, which is the winner of the LSMDC 2017 Text-to-Video and Video-to-Text retrieval challenge. Besides, in Table \ref{tab:sota-MPII}, we also investigate the effect of multi-modal fusion in our proposed TCE. Specifically, when we just use the 2048-D globally-pooled appearance features to describe the video, our model still outperforms most of the listed methods in Table \ref{tab:sota-MPII}. By augmenting the video representation with the motion feature, we can obtain a relative improvement of 25\% w.r.t. R@10. The audio features can further stably improve the performance. That is to say,  TCE has the potential of improving its performance on MSR-VTT by leveraging more informative features. Since the multi-modal fusion is not our focus in this paper, we leave the fusion experiment on MSR-VTT for future study.

\subsubsection{Ablation Studies}
To effectively answer the research question \textbf{R2}, we conduct extensive ablation studies on MSR-VTT based on the standard split. Specifically, we mainly organize the ablation studies into two groups: one for query encoder and the other for video encoder. The batch normalization is used to normalize the query/video representation in the following counterparts. \\
\noindent{\textbf{\underline{On Query Encoder: }}} We use the following baselines and variants to transform the natural language queries to vector representations. 
\begin{itemize}
	\item \textbf{WordEmb+AvgP} and \textbf{WordEmb+MaxP}: Add a fully connected (FC) layer after the word embedding layer and aggregate its output with average-pooling (AvgP) operation or max-pooling operation (MaxP).
	\item \textbf{LSTM} and \textbf{LSTM+AvgP}: Instead of composing the latent semantic tree, we directly use the last hidden state (LSTM) or apply an average-pooling over the output of LSTM (LSTM+AP).
	\item \textbf{TCE (w/o-Cxt)}: Remove the memory-augmented context vector $\mathbf{u}_t$ in the score module (Eq. (\ref{score})) and directly normalize the scaled dot-product between a global query vector and the hidden state of nodes. It is the standard implementation in \cite{choi2018learning}.
	\item \textbf{TCE (w/o-LSTM)}: Remove the leaf node LSTM module. Instead, we use an FC layer to transform the word embedding to the default input of TreeLSTM.
	\item \textbf{TCE (w/o-TAtt)+AvgP}: Remove the text attention module in Eq. (\ref{queryatt}), instead, we use the average-pooling operation.
\end{itemize}
\noindent{\textbf{\underline{On Video Encoder:} }} We use the following baselines and variants to transform videos to vector representations.
\begin{itemize}
   \item \textbf{Frame+AvgP} and \textbf{Frame+MaxP}: Use a FC layer to first transform the frame features, followed by average-pooling (AvgP) or max-pooling (MaxP).
   \item \textbf{GRU} and \textbf{GRU+AvgP}: Directly use the last hidden state of GRU or apply average-pooling over the output of GRU (GRU+AvgP).
   \item \textbf{TCE (w/o-Mha)}: Remove the multihead attention module.
   \item \textbf{TCE (w/o-GRU)}: Replace the GRU module with a FC layer to transform the frame embedding.
   \item \textbf{TCE (w/o-VAtt)+AvgP}: Remove the temporal video attention module in Eq. (\ref{vidatt}) with an average-pooling operation instead.	
\end{itemize}
Note that in the ablation studies, we only change one (e.g., query) part of our proposed TCE to the above baselines or variants, while keeping the rest of TCE (e.g., video) unchanged. 

Table \ref{tab:ablation-msrvtt} shows the performance comparison of our proposed full TCE model with different ablations on the MSR-VTT dataset. 
\begin{itemize}
\item Overall, we observe that our full model performs the best except in terms of MedR. Removing each component from TCE, such as \textit{Cxt}, \textit{Mha}, \textit{LSTM/GRU}, and \textit{TAtt/VAtt}, would result in relative performance degeneration, but not dramatically. It not only reflects the effectiveness of each component of our TCE, but also shows the robustness of our method. Each module can effectively complement each other, but is not very sensitive to each other.
\item There are also some interesting findings: the RNNs do not play a much more important role than we wish in this task. Compared with the baselines WordEmb+AvgP and Frame+AvgP, the LSTM and GRU help to improve the accuracy by a small margin, due to the modeling of the dependence between words/frames. However, if we remove LSTM or GRU from our query encoder or video encoder, the model exhibits a minor performance degenerates. TCE (w/o-LSTM) and TCE (w/o-GRU) still report high accuracy. This indicates the effectiveness of our latent semantic tree in capturing the structure information of the complex queries. It also reveals the necessity of the temporal interaction module that models the frame-wise feature interaction beyond the dependence between consecutive frames.
\item We also observe that the widely used average-pooling strategy does not performs well for the complex-query video retrieval task. Our introduced attention mechanisms in Eq. (\ref{queryatt}) and (\ref{vidatt}) performs well by attending to the informative constituent nodes and frames. 
\end{itemize}

\begin{table} [tb!]
	\renewcommand{\arraystretch}{1.0}
	\caption{Ablation studies on the MSR-VTT dataset using the standard dataset split \cite{xu2016msr} to investigate the effects of the tree-based query encoder and the temporal-attentive video encoder. The proposed method performs the best.}
	\label{tab:ablation-msrvtt}
		\vspace{-0.12in}
	\centering 
	\scalebox{0.95}{
		\begin{tabular}{|l|c|c|c|c|}
			\hline
			\textbf{Method}   & \textbf{R@1} & \textbf{R@5} &\textbf{ R@10} & \textbf{MedR} \\
			\hline\hline
			\textit{\textbf{On Query Encoder}} &  & & &  \\
			WordEmb+AvgP       & 6.79 & 20.98 &30.68 & 32 \\
			WordEmb+MaxP        &5.92& 18.90 &27.82 & 40  \\
			LSTM                   & 6.91 &21.31 & 31.17 & 31 \\
			LSTM+AvgP       & 6.95 & 21.28 & 30.68 & 35 \\
			\hline
			TCE (w/o-Cxt)  & 6.98 & 21.46 & 31.49 & 30  \\
			TCE (w/o-LSTM) & 7.09 &21.86 &31.67 & 31 \\
			TCE (w/o-TAtt)+AvgP  &6.59 & 20.57 & 30.48 & 34  \\
			\hline\hline
			\textit{\textbf{On Video Encoder}} &  & & &  \\
			Frame+AvgP     & 6.67& 20.41 & 29.89 & 36 \\
			Frame+MaxP   &6.20 & 20.24 & 29.87 & 35 \\
			GRU                   & 6.75 & 21.03 & 30.91 & 31  \\
			GRU+AvgP       & 6.17& 19.51 & 28.71 & 38 \\
			\hline
			TCE (w/o-Mha)  & 6.97 & 21.59 & 31.19 & 31  \\
			TCE (w/o-GRU)  &7.08  &21.96 &31.86 & 30 \\
			TCE (w/o-VAtt)+AvgP  & 6.73 & 21.38 & 31.74& \textbf{29} \\
			\hline\hline
			\textbf{TCE}  & \textbf{7.16} & \textbf{21.96} & \textbf{32.04} & 30  \\
			\hline
	\end{tabular}}
	\vspace{-0.15in}
\end{table}
\begin{figure}[tbp]
	\centering
	\subfloat[Grouped by query lengths.\label{fig:query_length}]{
		\includegraphics[width=0.96\columnwidth]{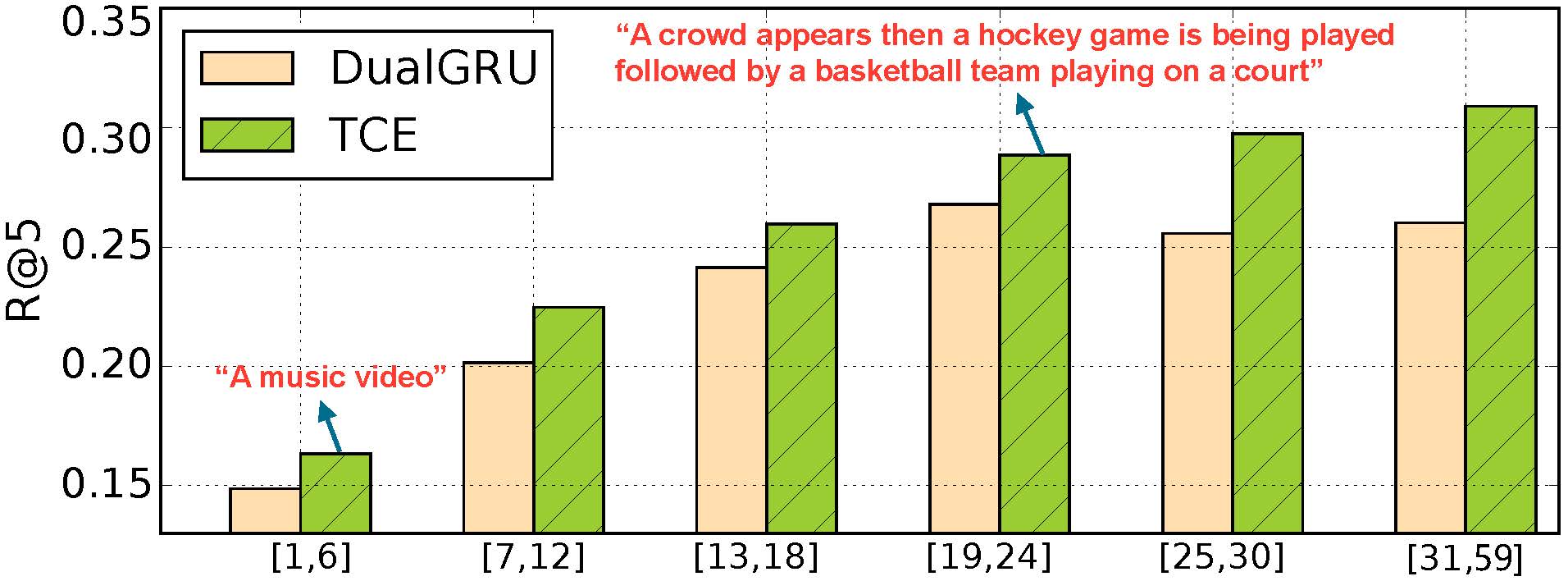}}\\
		\vspace{-0.15in}
	\subfloat[Grouped by query categories.\label{fig:category_perf_group}]{
		\includegraphics[width=0.96\columnwidth]{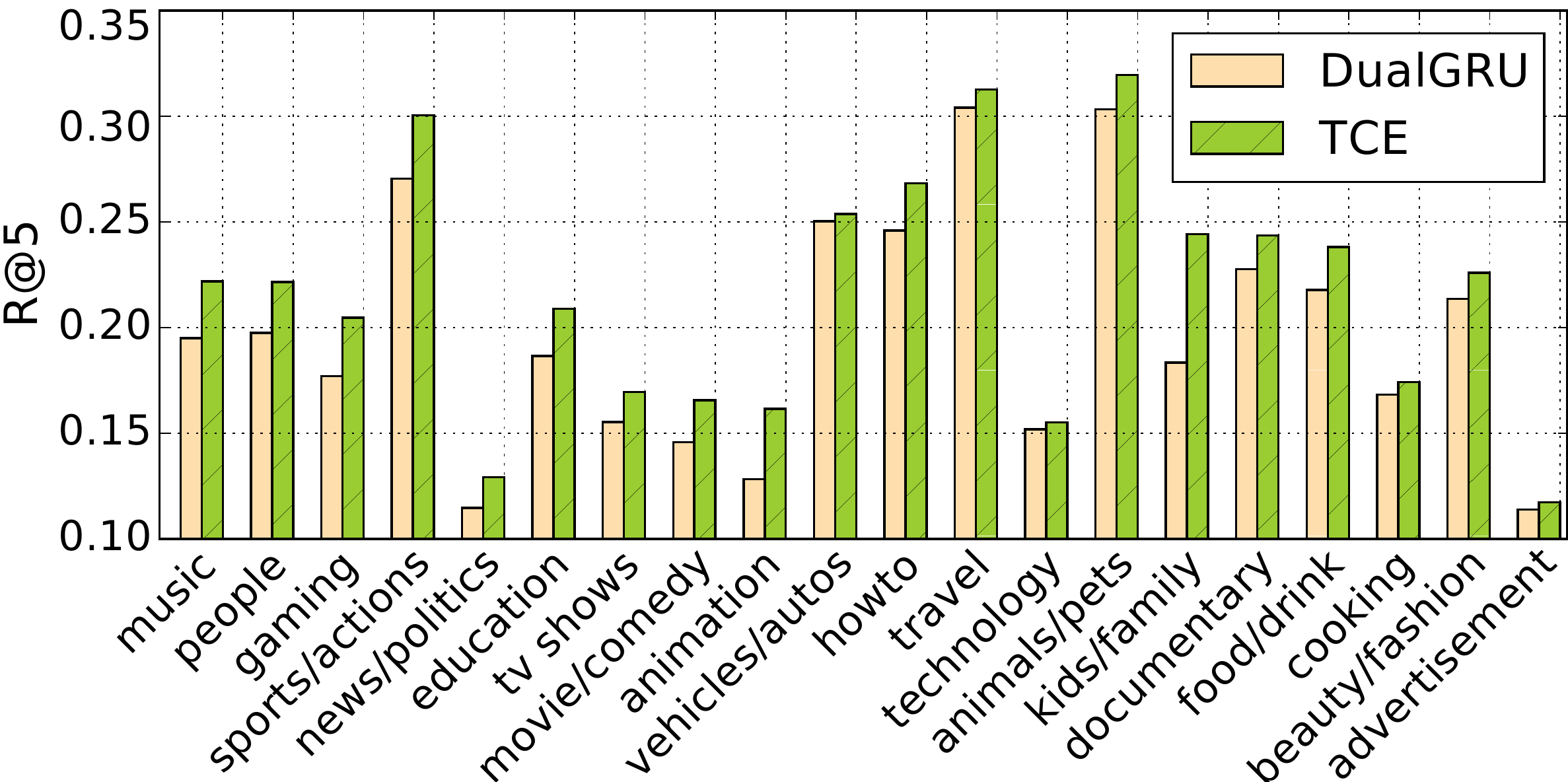}}
		\vspace{-0.1in}
	\caption{Performance comparison of DualGRU and our proposed TCE on MSR-VTT. Queries have been grouped in terms of (a) query lengths and  (b) query categories.}
	\label{fig:perf_group}
	\vspace{-0.15in}
\end{figure}

\begin{figure*}[htbp]
	\centering\includegraphics[width=2\columnwidth]{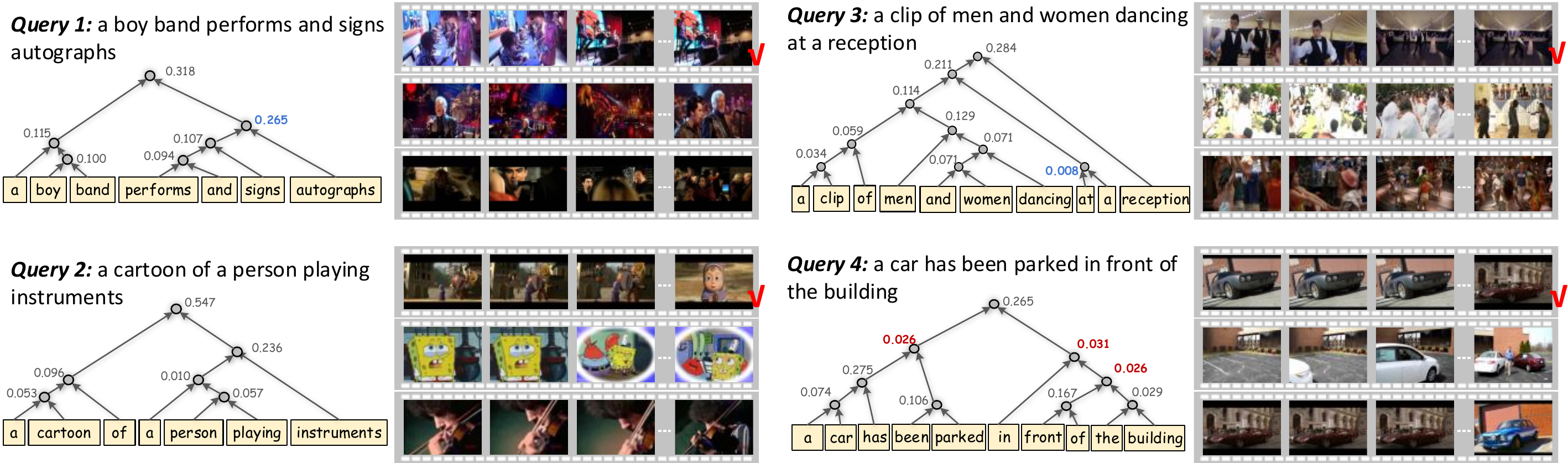}
	\vspace{-0.1in}
	\caption{Four examples obtained by our TCE model on MSR-VTT. The composed latent semantic trees are presented under the corresponding queries. The normalized node weights are also shown. Videos with red marks are the correct ones. }\label{fig:retrieval_result}
	\vspace{-0.1in}
\end{figure*}
\subsubsection{Analysis on Different Types of Queries.(\textbf{R3})}
To investigate how our proposed TCE perform on different groups of complex queries, we group 59,800 test queries of the MSR-VTT dataset (Data split from \cite{xu2016msr}) according to their query lengths and categories. We compare TCE on different groups with the baseline model DualGRU which utilizes the bidirectional GRU with average pooling for both text encoding and video encoding.
The performance comparison in each group by query lengths and categories are shown in Figure \ref{fig:perf_group}(a) and \ref{fig:perf_group}(b), respectively. 
In Figure \ref{fig:perf_group}(a), our proposed TCE consistently outperforms the DualGRU in all groups with different query lengths, showing its effectiveness in complex query modeling. Especially, we clearly observe that with increasing query lengths (from left to right in Figure \ref{fig:perf_group}(a)), the performance gain of TCE over DualGRU becomes much more significant.
Generally, the longer queries are more complex than the shorter queries. As demonstrated in Figure \ref{fig:perf_group}(a), the query ``\textit{a crowd appears then a hockey game is being played followed by a basketball team playing on a court}'' is more complex than the query ``\textit{a music video}''. Hence, the results validates that our proposed TCE is better in handling the complex queries, mainly benefiting from the latent semantic tree. 
In Figure \ref{fig:perf_group}(b), the performance in different query categories varies greatly, showing the varying difficulty of queries in different categories.
For instance, the performance on the query group of \textit{sports/actions} is higher than 0.25, while only about 0.13 on the query group of \textit{news/politics}. The observation is reasonable since the \textit{sports}/\textit{actions} scenes are easy to be visually distinguished, while the \textit{news/politics} scenes are much more diverse, and it is hard to learn the relation between news words and the visual scenes with limited training data.
Despite the varied difficulty for each group, our proposed TCE model consistently beats the baseline on all groups.

\subsubsection{Qualitative Analysis}
Figure \ref{fig:retrieval_result} shows four cases of qualitative results about the composed tree structures and the retrieved videos. For each query, the top three videos retrieved from the MSR-VTT are showed. Although only one correct video is annotated for each query, the retrieved three videos in Figure \ref{fig:retrieval_result} are typically semantically relevant to the given query to some extent, showing the effectiveness of TCE. We observe that our approach is able to construct syntactically reasonable tree structures (\eg \textit{Query 1} and \textit{Query 2}) and also identify the informative constituent nodes based on the attention mechanism, thus being helpful to better understand the complex query. For example, in \textit{Query 1}, ``\textit{performs and signs autographs}'' describes the \textit{action} of the video clip, which is easy to be visually distinguished and usually reflects the main search intention, whose corresponding node in the tree was assigned a relatively large weight of 0.265. In \textit{Query 2}, ``\textit{a person playing instruments}'' refers to the key search intention, whose corresponding node in the tree was assigned a relatively large weight of 0.236. For \textit{Query 3}, the composed tree is far from perfect, while the node ``\textit{at a}'' contains less semantic information and was reasonably assigned the smallest attention weight of 0.008. For \textit{Query 4}, the composed tree is syntactically reasonable, but some relative important nodes were assigned with small attention weights. Although some noises have been introduced in the latent semantic tree construction, our model still finds relevant videos for \textit{Query 4}, which shows the robustness of our model.

\section{Conclusion}
In this work, we proposed a novel framework for complex-query video retrieval, which consists of a tree-based complex query encoder and a temporal attentive video encoder. Specifically, it first automatically composes a latent semantic tree from words to model the user query based on a memory-augmented node scoring and selection strategy and then encodes the tree into a structure-aware query representation based on an attention mechanism. Besides, it jointly models the temporal dependence between frames and frame-wise temporal interaction in the temporal attentive video encoder, followed by an attentive pooling mechanism to vectorize the video. 
This work provides a novel direction for complex-query video retrieval by automatically transforming the complex query into an easy-to-interpret structure without any syntactic rules and annotations. In the future, we will explore the proposed approach for other language-guided video tasks, such as video moment retrieval with natural language~\cite{liu2018cross}. We also plan to integrate the multimedia indexing technique~\cite{hong2017coherent} with our approach for large-scale retrieval. We are also interested in exploring the external knowledge to enhance the text representation learning and the tree construction~\cite{cao2017bridge,cao2018joint} in the future study.

\section{Acknowledgments}
This research is supported by The National Key Research and Development Program of China under grant 2018YFB0804205, the National Research Foundation, Singapore under its International Research Centres in Singapore Funding Initiative, the National Natural Science Foundation of China under grant 61902347, and the Zhejiang Provincial Natural Science Foundation under grant LQ19F020002. Any opinions, findings and conclusions or recommendations expressed in this material are those of the author(s) and do not reflect the views of National Research Foundation, Singapore.


\bibliographystyle{ACM-Reference-Format}
\bibliography{sigir2020}

\end{document}